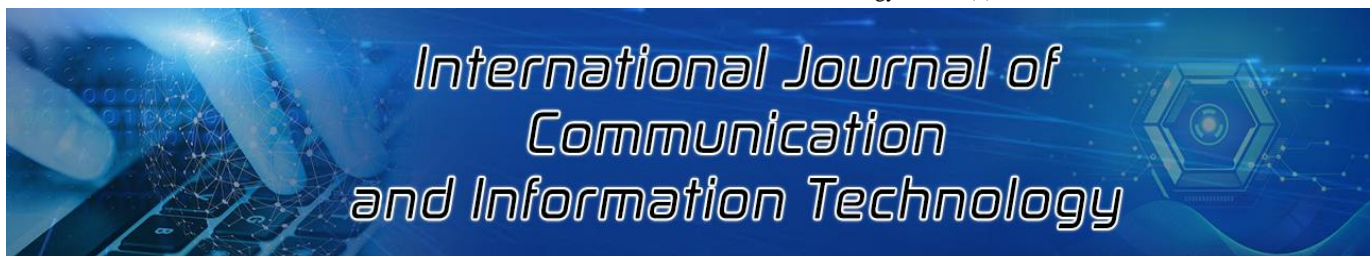

**Aya Kaysan Bahjat**
Informatics Institute for Postgraduate, Studies, Baghdad, Iraq

# A survey of facial recognition techniques

**Aya Kaysan Bahjat**

**Abstract**
As multimedia content is quickly growing, the field of facial recognition has become one of the major research fields, particularly in the recent years. The most problematic area to researchers in image processing and computer vision is the human face which is a complex object with myriads of distinctive features that can be used to identify the face. The survey of this survey is particularly focused on most challenging facial characteristics, including differences in the light, ageing, variation in poses, partial occlusion, and facial expression and presents methodological solutions. The factors, therefore, are inevitable in the creation of effective facial recognition mechanisms used on facial images. This paper reviews the most sophisticated methods of facial detection which are Hidden Markov Models, Principal Component Analysis (PCA), Elastic Cluster Plot Matching, Support Vector Machine (SVM), Gabor Waves, Artificial Neural Networks (ANN), Eigenfaces, Independent Component Analysis (ICA), and 3D Morphable Model. Alongside the works mentioned above, we have also analyzed the images of a number of facial databases, namely JAFEE, FEI, Yale, LFW, AT&T (then called ORL), and AR (created by Martinez and Benavente), to analyze the results. However, this survey is aimed at giving a thorough literature review of face recognition, and its applications, and some experimental results are provided at the end after a detailed discussion.

**Keywords:** Face recognition, aging, illuminations, pose variation, partial occlusion

## 1. Introduction
The twenty-first century is generally accepted as an era of remarkable scientific advancement, whereby significant gains have been made in the swift empowerment of individuals geared towards the achievement of their goals. To reinforce this claim, the modern day adoption of computer technology is now an inseparable part of life, which can hardly be done without in the modern world. The use of computers has been spread in a very broad range of activities, including simple and more complicated activities like problem solving. One of such contributions is the facial-recognition technology, which is a valuable tool when it comes to identifying facial features with the use of intrinsic characteristics. It has turned into one of the best studied fields in computer vision and pattern recognition. Nevertheless, its relevance can be seen in the popularity of facial recognition in numerous applications, which include biometrics, surveillance and police access control, information security and smart cards. Nonetheless, there are a number of issues that researchers are facing which include partial occlusion, aging, pose variance, illumination variability, facial expression.

The face recognition concept can be divided into two phases, namely, face verification and face identification. The system in the former stage matches a given facial structure in a given image to a particular individual and in the latter stage the salient facial features are extracted and compared with ones coded in a facial data repository to determine who the person in the image is. However, some of the current authentication procedures that are based on face recognition have a lack of reliability. As an example, traditional security relationships like wallets, smart cards, tokens, and cryptographic keys often use passwords and personal identity numbers that by design exert a heavy cognitive burden on the user making them hard to remember. Moreover, it is easy to forget these codes and passwords; moreover, these magnetic cards may be lost, steal, or copied. They consequently become illegible. They cannot be lost, forgotten, or misplaced as it is the case with the biological traits and characteristics of a person [1]. Biometric recognition system may be assembled through a number of methods. However, the most used modalities include the iris and the fingertips. These modalities involve the intervention of the individual to get to the system.

**Corresponding Author:**
**Aya Kaysan Bahjat**
Informatics Institute for Postgraduate, Studies, Baghdad, Iraq

Additionally, more recent systems have access by the participants without direct intervention. Face recognition are among the most possible technologies that are capable of easily tracing and watch an individual within the system. The face recognition databases that are controllable up to the non-controllable videos, such as the use of (YouTube Faces) YTF [2] in videos and (Labeled Faces in the Wild) LFW [3] in pictures.

## 2. Databases
There are numerous databases of faces public and private to be used in research. These databases are distinguished amongst each other based on several factors. The most significant are on the following:
- The most critical standard is the number of photos in each database.
- Photo count of each distinct class: since each individual has a characterization in a class c, the count of photos in a particular class represents the count of representative photos of the individual. Actually, the photographs are captured in different circumstances (Aging, facial expression...etc.).
- Presence of an occlusion (beards, glasses, etc.).
- Gender of the obtained persons.
- The volume of the pictures.
- The presence of immobile images or videos.
- The time among recordings.
- Face alignments and pose.
- The changing in lighting.
- The presence of a common background.

It is, therefore, recommended to choose the correct database when appraising the algorithm. Practically, various databases have clearly defined protocols that can be used in comparison of results directly. Additionally, the decision has to be informed by the research question that is being addressed, such as aging, facial expression, illumination, and so forth. The fact that several pictures are given to one person can also be a key aspect of proper functionality of the algorithm. As shown in Table 1, the extensive collection of 2D face databases have different differences in terms of grayscale image versus RGB image, number of images of subject, total image count of a subject, and differences in images which include; occlusions (o), aging (a), pose (p), facial expression (e) and illumination (i).

**Table 1:** The Main 2D Face Databases (The picture differences are represent through (e) facial expression, (p) pose, (a) aging, (i) illumination and (o) occlusion)

| Database | RGB Color/grey | Number of Pictures/Persons | No. of Persons | Pictures Size | Variation |
|---|---|---|---|---|---|
| Yale [4] | Gray | 165 | 15 (14 men and 1 woman) | 320 × 243 | i,e |
| AT&T [5] | Gray | 10 (pictures per person) | 40 | 92 × 112 | i,a |
| XM2VTS [6] | RGB | 2360 | 295 | 576 × 720 | p |
| AR database [7] | RGB | 4,000 | 126 (56 women and 70 men) | 576 × 768 | o,e,i,a |
| CVL [8] | RGB | 7 (pictures per person) | 114 (6 women and 108 men) | 640 × 480 | e,p |
| Oulu Physics [9] | RGB | 16 for every person (and additional 16 if have glasses) | 125 | 428x569 | i |
| JAFEE [10] | Gray | 213 | 10 | 256 × 256 | e |
| FEI data-Base [11] | RGB | 2800 | 200 | 640x480 | p,e,i |
| LFW [3] | RGB | 13,233 | 5749 | 250 x 250 | P,i,e,o,a |

## 3. Structure of Face Recognition System
A develop face recognition system employs three major stages; 1) Face Detection 2) Feature Extraction 3) Face Recognition [12]. The stage of face detection is applied in the process of identifying and searching the person face picture obtained by the system. And feature extraction stage is employed to extract feature vectors to any person face detected in first stage and finally is the face recognition where features extracted in the person face are compared with all model face databases to establish the identity of person face. The Figure 1 shows the structure of the face recognition.

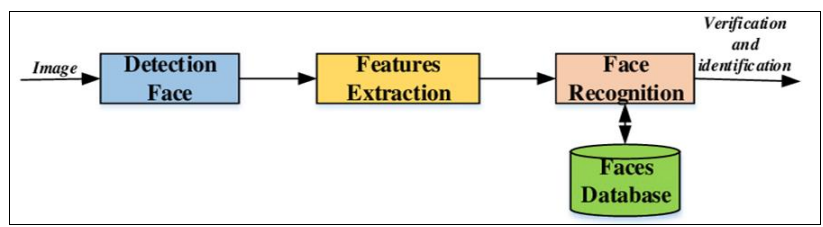

**Fig 1:** Face Recognition Structure [12]

### 3.1. Face Detection
The initial stage of face recognition system involves the location of human faces within a given picture. As shown in the Figure 2. The intent of this step is to establish whether there are faces of persons in the input picture or not. Facial rotation, lighting and...etc. can inhibit proper facial recognition. Preprocessing steps are used to simplify the development of face recognition system and make it even more powerful. Many methods are employed in detect and locate the picture of a person face, such as: Histogram Oriented Gradient (HOG) [13], Viola-Jones Detector [14] and Principal Component Analysis (PCA) [15] or face detection step can be applied to object detection [16], region of interest detection [16], picture and video classification etc.

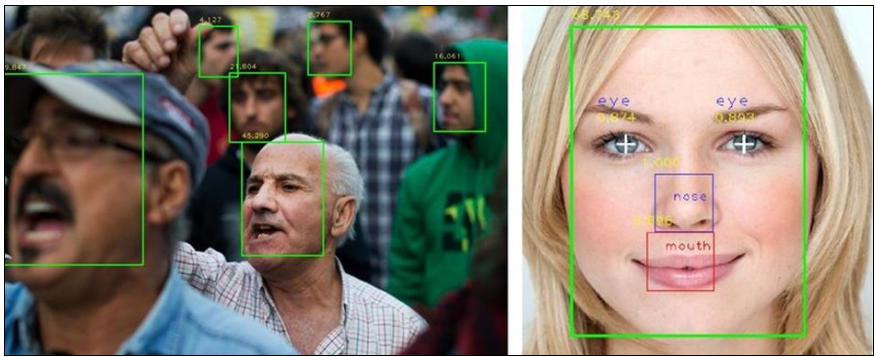

**Fig 2:** Example of the Face Detection [17]

## 3.2. Extraction of Features
The current step is critical in retrieving the same facial features that were found in the previous detecting stage. In the case of a face, as Figure 3 shows, such a step encodes a face into a collection of vectorial data, otherwise known as a signature into which we encode the salient features of the facial image otherwise known as the eyes, mouth and nose and their geometric projections [22, 14]. The size, shape, and structural arrangement of a face are distinctively defined and, therefore, allow recognition of a face. Majority of the modern methods hence focus on the extraction of a geometric form of the nose, mouth and eyes in order to determine the subject, normally through distance metrics and dimensional analysis [12] for instance: Local Binary Pattern (LBP) [18], Fourier Transforms [19], Local Phase Quantization (LPQ) [20], Scale-Invariant Feature Transform (SIFT) [13], Linear Discriminant Analysis (LDA) [21], Eigenface [21], and Histogram Oriented Gradient (HOG) [13] are wide used for extract facial features.

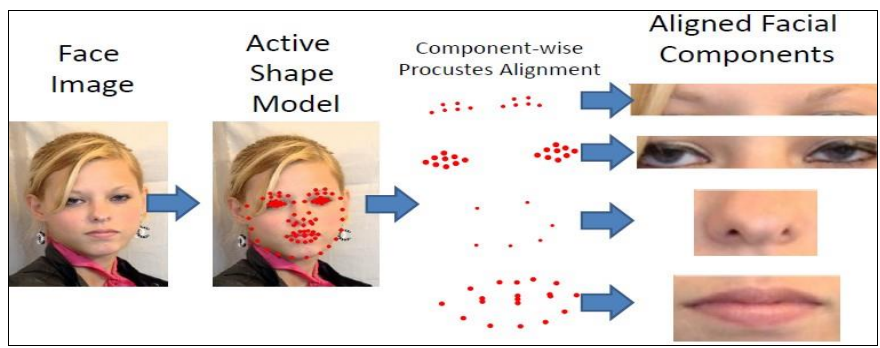

**Fig 3:** Face Component [22]

## 3.3. Face Recognition
This is done by extracting features of the background in the feature extraction stage which is then compared with the known face templates stored in a predetermined database. Face recognition generally occurs in two phases, namely, identification, which is followed by verification. In the process of identification, the test face is compared with a set of facial images to isolate the most similar prototype. Afterward, verification is done whereby the candidate face is compared to a given known template in the repository hence the decision to accept or reject is made [23]. State-of-the-art algorithms like k-nearest neighbors (K -NN) [24], convolutional neural networks (CNN) [25], and correlation filters (CF) [26] have shown good performance with regard to this task. Figure 4 shows the schematic expression of this process of recognition.

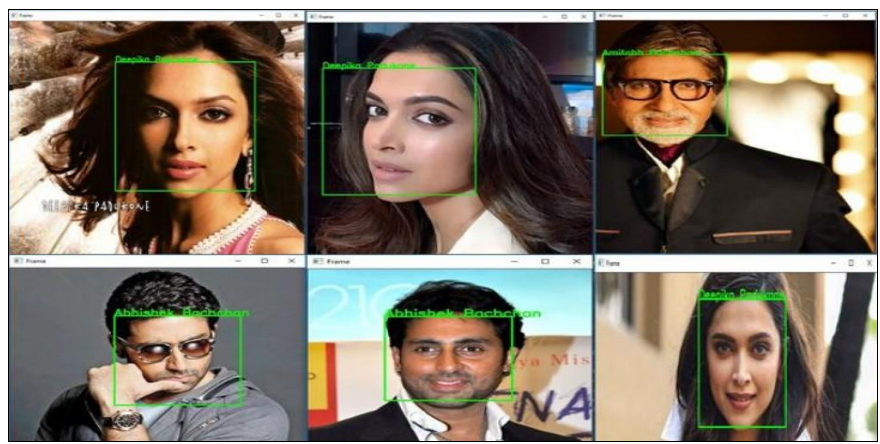

**Fig 4:** Illustrate the Face Recognition [27]

## 4. Models of Face Recognition
Face recognition applications use three different methods, namely 2D, 3D and video. Will will discuss the face recognition concerns in the following departments.

### 4.1. Face Recognition in 2D
The use of face recognition has been a research area over ten years using 2D still images [28]. Facial recognition systems relying on 2D fixed picture entail the process of capturing a user and comparing it to a collection of captured images to recognize a person. Under this technique, the user is expected to cooperate and possess a frontal picture of face with plain background under the same lighting conditions to facilitate the acquisition and division of facial picture of high quality. Nonetheless, it is now understood that small deficiencies of illumination and pose can have a drastic effect in lowering the performance of face recognition systems based on a single-shot 2D picture [29]. 2D face recognition is typically categorized by the amount of pictures per picture used as illustrated in Table 2.

**Table 2:** Face Recognition Scenarios in 2D Domain

| Probe | Gallery | |
|---|---|---|
| | Single still picture | Many still pictures |
| Single still picture | One-to-one | One-to-many |
| Many still pictures | Many-to-one | Many-to-many |

A number of highly developed 2D face recognition algorithms rely on Linear Discriminant Analysis (LDA) [30], Correlation- Based Matching [31], the Elastic Graph Model (EGBM) [32] and Principal Component Analysis (PCA) [33].

### 4.2. Face Recognition in 3D:
The methods of 3D face recognition make use of the shape of the surface of the face [34]. Unlike 2D face recognition, 3D methods are resistant to changes in illumination and pose due to the existing invariance of a three-dimensional shape to these criteria. The depth sensor used to capture the 3D picture of the face generally covers around 120o right-left. This half-dimensional image is known as a 2.5D image. Noting that one can sum 3-5 of these 2.5D scans together, one can build a complete 3-D model of a full 360 degree view of the face. As a rule, the probe is a 2.5D capture, and the database can have either 2.5D records or an entire 3-D reconstructed model. It is possible to do either 2-depth image-to-2-depth image identification [34] or 2-D image-to-3-D face model identification [35]. Table 3 builds on Table 2 as it includes 2-D and 3-D face representations.

**Table 3:** Face Recognition Scenarios across 2D and 3D Domain

| Probe | Gallery | |
|---|---|---|
| | 2D Pictures | 3D Models or 2.5D Pictures |
| 2D Pictures | 2D to 2D | 2D to 3D |
| 2.5 Pictures | 3D to 2D | 3D to 3D |

It has a lot of ways based on three-dimensional models which have been computed based on a series of two-dimensional images [36]. The resulting 3D model is used to produce a series of two dimensional projections, which are reflective of the probe images [35]. Moreover, the reconstructed 3-d image can generate a direct view of the probe taking arbitrary pose and light conditions. This synthesized probe image is then identified by the comparison with the frontal counterpart. Figure 5 depicts a three dimensional model of the face alongside the accompanying two dimensional projections of the face taken in diverse poses and light-reflectance scenarios [37].

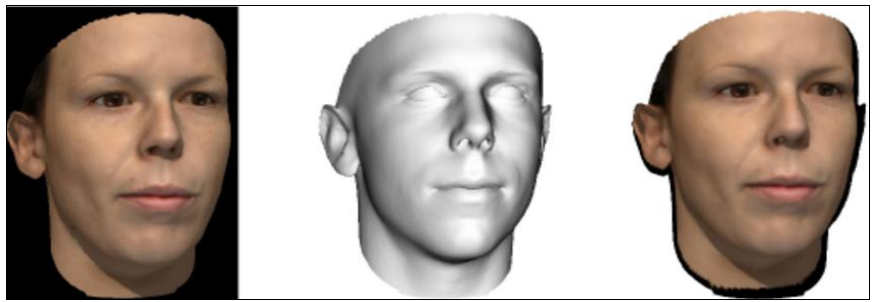

**Fig 5:** A 3D face Model and its 2D Projections [38]

### 4.3. Face Recognition in Video
Although traditional face-recognition systems mostly use still images, there is an emerging academic trend of creating robust face-recognition systems that would use video as input. Face recognition has been a subject of interest through video based recognition because of the prevalence of surveillance cameras. The capability to recognize faces automatically in real time with this type of footage would be applied in a variety of ways such as secret identification of a person through an established network of surveillance cameras. However, the images of the faces obtained by video streams belong to off-frontal postures and the illumination differences are usually high, which harms the work of many commercial face-recognition systems.

Two conspicuous features of the video data include: i) it provides multiple frames of the same object, and, iii) it is provided with the time. There is variation in multiple frames which enables one to select a high quality frame like a good frontal face photograph that will give a better recognition performance. The changes of the face are dynamic and they are related to the information which is lived in time in the video. Nevertheless, the question whether the face motion is able to encode identity-related cues is difficult to determine; more studies are needed to be able to utilize the time dimension. The use of video capabilities can contribute to the work of face-recognition systems. Figure 6 shows 4 frames of a perfect video shot of face-recognition study [37].

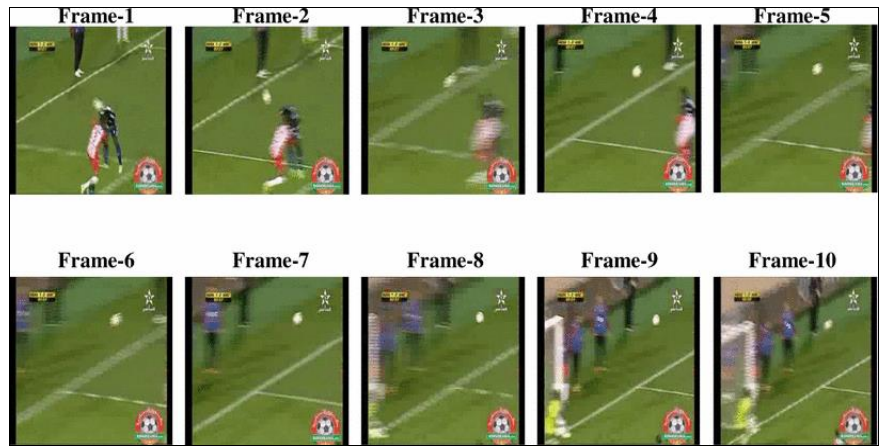

**Fig 6:** Four Frames from a Video [19]

## 5. Challenges and Difficulties
Although there has been a gradual advancement of a variety of face recognition algorithms, there are still many challenges involved. These issues still present serious setbacks to these algorithms as explained further below [39]:

### 5.1. Several databases of facial pictures, such as references
They are clustered to compare the facial recognition algorithms. All databases are maintained so as to analyse a certain factor which may be; occlusion, pose, facial expression or illumination.
Past investigations show that face recognition systems are already mature when there is controlled environment; however, when these are applied in the field, there are a lot of complexities.

### 5.2. Face recognition is special and difficult state from object recognition:
Facial recognition challenge is given the phenomenon of frontal imagery which is more likely to show the high level of visual similarities, and thus the difference variances are usually minute, and thus require rigorous scrutiny of analysis. Recent studies using the cutting-edge recognition algorithms, which have been used to test state-of-the-art repositories like FAT, FERET, and FRVT have identified the top barriers to algorithmic efficacy as illumination, age and pose.

### 5.3. The algorithms of face recognition are selected as desired by the application:
As an example, feature-based techniques cannot be used on low-resolution (e.g. 15 x 15 pixels and smaller) facial images. The other question is related to the correct usage of dimensionality-reduction methods: in what cases do we want to use Linear Discriminant Analysis, and in what cases do we want to use Principal Component Analysis or Independent Component Analysis when developing the system?

### 5.4. Implementation details usually specify the performance of the system:
The modern image processing practice is that the input visual data should be normalised in regards to in-plane rotations, masking, affine warping as well as scaling so that accurate alignment of the underlying shapes can be met.

### 5.5. Low resolution is a significant factor in face recognition when taking pictures from a far distance"
In addition, eye occlusion negatively impacts the accuracy of most facial recognition systems; which normally normalize and scale the input images then identify them.

### 5.6. When the face occluded, the recognition rate drops rapidly:
Similarly, the structural elements that significantly affect the recognition rate are mustaches, glasses, and beards.

### 5.7. For good recognition performance, accurate feature location is critical:
Upon viewing a face under a specific angle, some of its constituent features are deformed thereby rendering ineffective various facial recognition algorithms. The major confounders that are still challenging the contemporary face recognition algorithms are age, lighting, and pose.

## 6. Factors Affecting of Face Recognition
Detecting faces of human beings on photographs and video sequences is a very difficult issue. Although numerous methodological theories have been suggested, none of them can be absolutely accurate, which is mostly explained by numerous issues that are inherent to the problem area.
These complications may be classified into two major groups namely, internal and external factors as outlined by [40].
Internal factors include intrinsic qualities like the physiological state of the face, expressive dynamics and age-induced changes, which affect system performance. External factors on the other hand are pertinent to the environment, which change the look of the face such as changes in light.

### 6.1. Aging
The problem of aging is considered one of the most important confounding factors of modern face-recognition techniques, and it actually makes biometric verification a cluttered, noisy procedure to most algorithms. This need of stability is the fundamental feature of any biological measure which is to be considered as a credible biometric signal. The human face is a heterogeneous complex of epidermal tissue, bones, and moving muscles; any tightening of the latter is bound to cause micro-deformations of the extrinsic landmarks of the face. However, these properties become increasingly wasted, which leads to significant changes in the appearance of an individual as he grows old-old, they occur most vividly in Figure 7: one notices the development of dermal textures, including fine wrinkling and progressive alteration of facial features. The longitudinal changes, therefore, have to be considered in any viable recognition system that is to be engineered. A body of study work that is directly directed to the reduction of age related drift has therefore been pursued [40]. Slow nature of biological ageing, though, poses an extreme constraint in terms of the size of longitudinal data that can be used to train strong classifiers and makes amassing age-progressive samples a daunting task. Accordingly, studies on age strength have received extensive academic interest over the last few years.

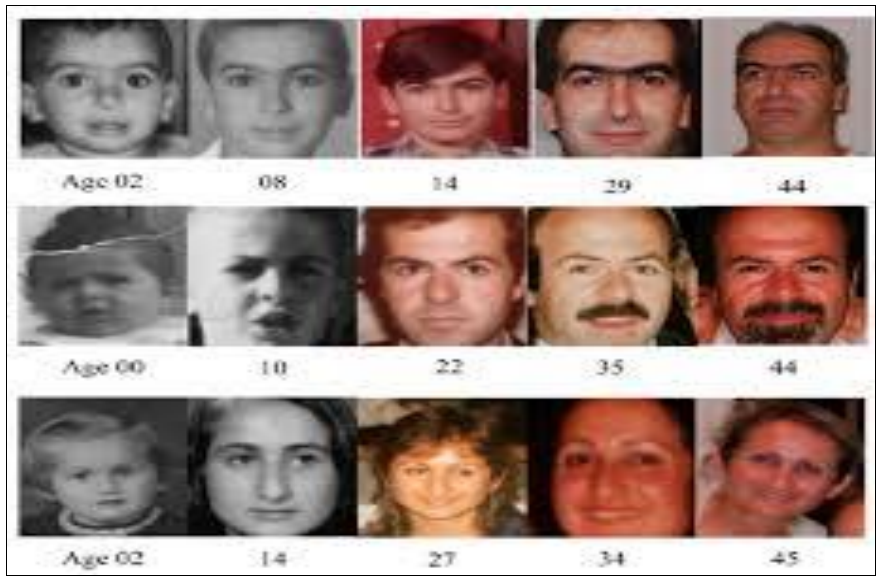

**Fig 7:** Pictures of the Same Subject at Age (a) 5, (b) 10, (c) 16, (d) 19 and (e) 29 [41]

**6.2. Facial Expression:** Facial expression is a non-verbal communication mode, which is supposed to pass messages by the movement of facial muscles. However, facial expression heterogeneity provides ambiguities to face-recognition systems. A number of face-recognition systems have been designed that work well on images of a controlled setting. The diverse expression of the face corresponds to different situations and moods, which change the facial geometry, as shown in Figure 8, and even slight changes in the image make it difficult to identify the face. The studies have been exploring face recognition keeping in mind facial expressions [42]. This issue can be solved by various methods including muscle-based, motion based and model-based methods.

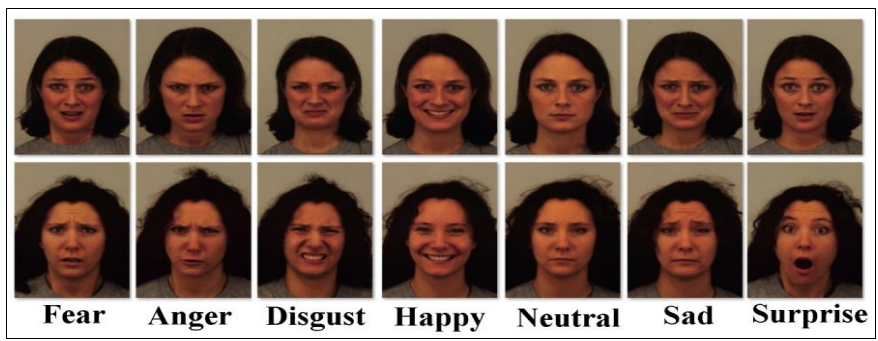

**Fig 8:** Examples of some Facial Expressions from a Daily Life [42]

**6.3. Pose Variation**
Varying head pose is a major impediment to the efficiency of the facial recognition systems. This is because when subjects are photographed, their pose is varied in every capture as is depicted in Figure 9. The lack of a canonical pose means that it is difficult to discriminate and identify faces in pictures with different poses. As a result, a lot of systems are limited to work under rigid imaging conditions.
The strategies that have been used to reduce pose variation can be categorized into two groups depending on the nature of the images in the gallery, which are pose-specific recognition and multi-view face recognition. The latter can be discussed as some sort of a frontal-face recognition.
The annex emphasizes the need to include the gallery images of every different pose. On the other hand, implementing the facial recognition, it is crucial to take into consideration faces which are not familiar to the system. A powerful face recognition method should therefore be able to withstand pose variations and still be able to identify a subject in various poses. There are still many challenges in this sphere. Indeed, there is a number of researchers who are currently working towards the resolution of these deficiencies. To this end, the ideal system that is accurate has not been realized. The problems of variable face recognition and pose change may be reduced with the help of several approach and methodologies that are usually grouped in three major categories: 2D techniques, general algorithms, and 3D models.

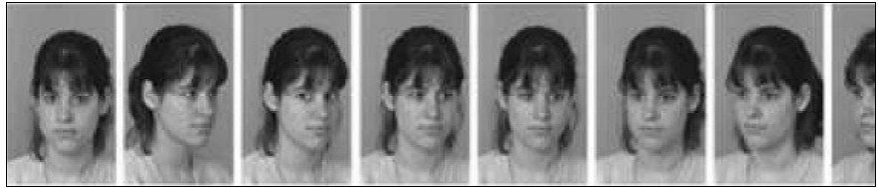

**Fig 9:** Examples of some Pose Variations from FERET Database [44]

**6.4. Partial Occlusion:** Occlusion can be defined as artificial or natural barriers which are found in an image.
The face-recognition methods that tackle the problems of partial occlusion can be broadly categorized into a few different groups, namely, Fractal-Based Methods as well as Feature-Based Methods, and Part-Based Methods [45]. The partial occliction has affected many parts of the face- ears are usually affected by earrings to a degree that the recognition of faces becomes difficult.
Occlusion has been shown to have a negative impact on system performance as people deliberately block their faces with scarfs, sunglasses, veils, or holding their mobile phones or hands in front of their faces as shown in Figures 10 and 11.

It has been noted sometimes that there are other factors, like the shadows created by strong lighting, which are occlusive factors. On the issue of faces which are partially covered, we will go ahead and use the strategies of analysis. Such means of analysis break down the face into various parts. The issue could be avoided by the removal of some of the features, which result in the imprecise recognition of the face in the picture. Often these methods of analysis rely on the analysis of features, in which the most informative features are identified and then aggregated. Another direction is a nearly holistic one, where features which are occluded are disregarded and the remaining facial data is used as useful data. Scientists are now coming up with numerous methods of overcoming this problem [46].

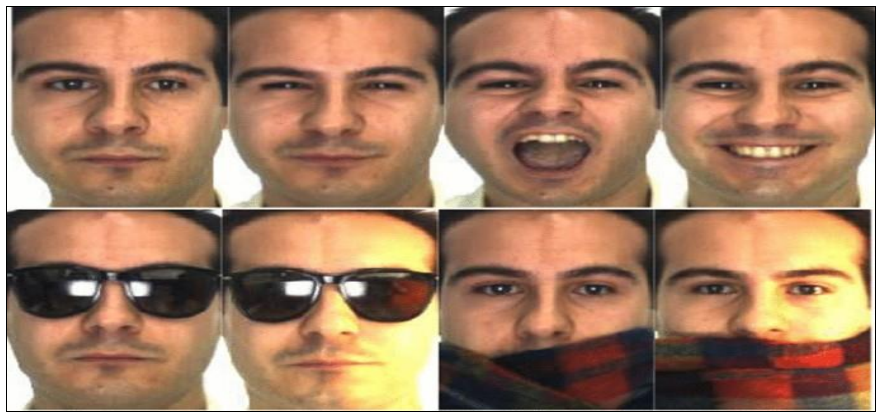

**Fig 10:** Example of the Partial Occlusion [21]

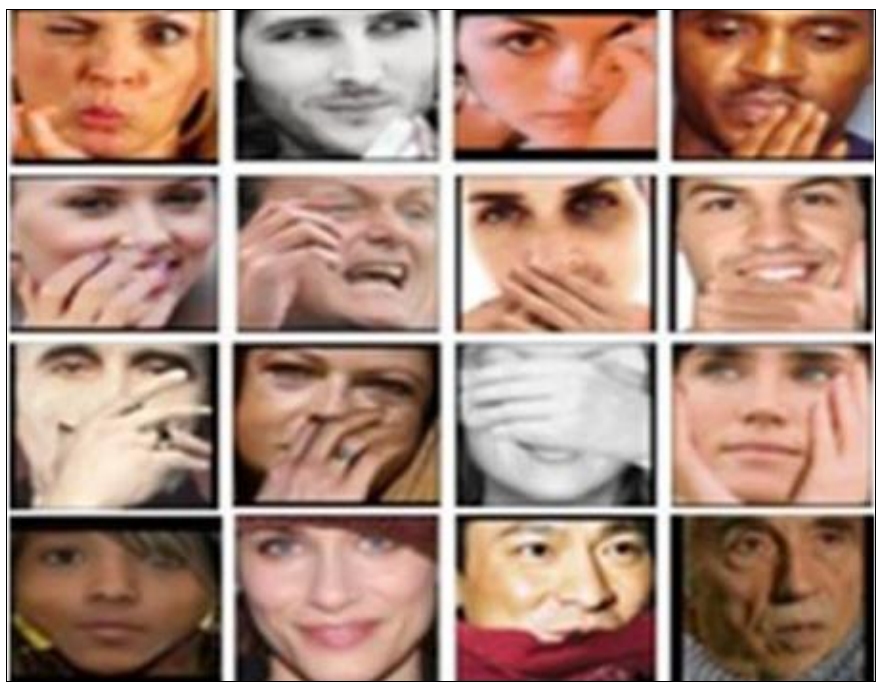

**Fig 11:** Faces with Occlusion Effect as Presented in Hand2Face [47]

**Ibn Al-Haitham Jour. for Pure & Appl. Sci. 34 (2) 2021**

**6.5. Effect of Illumination**
Changes in illumination significantly affect face recognitions, as shown in Figure 12, and as such, have attracted a lot of scholarly interest.

The accurate recognition of an individual or individuals on the basis of still photos or video recordings has been getting harder to attain. Whereas it is relatively easy to extract the desired information of photographs taken in controlled conditions, where the background is homogenous, the recognition of faces taken in uncontrolled ones, with heterogeneous backgrounds, is quite a challenge. Such complexity is due to shadows, under exposure and over exposure among other factors.

Special attention has been paid by researchers to the resolution of the given problems, and methods have been devised that alleviate the changing illumination effects and enhance the capability of recognition.

There are three major paths of dealing with this problem: face-reflection-field estimation, gray-level normalization and gradient-based edge extraction. The gray level conversion algorithm makes a pointwise mapping either by means of linear or nonlinear functions. The edges are extracted in the gray-level image using gradient based techniques. The fact that illumination is also an important confounding variable in face recognition in video or still images means that the above methods have been optimized to reduce the negative effect brought about by the issue [48].

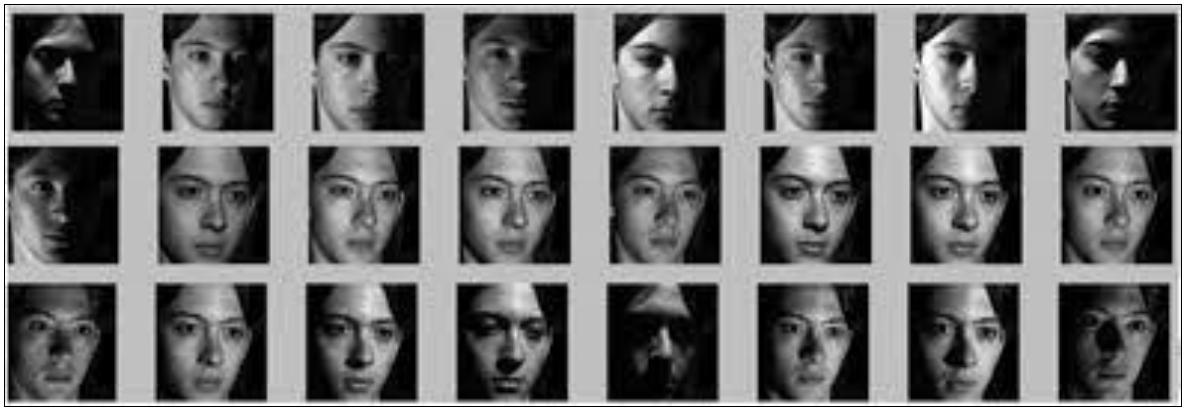

**Fig 12:** Example of some Illumination Variation from Yale Database b [49]

**Advantages of Face Recognition**
**High Success Rate:** The current facial biometric technology has shown a significantly high level of accuracy particularly with the introduction of three dimensional facial recognition system. Such systems are so much protected against fraudulent activities such that the stakeholders can trust such systems to effectively record attendance and time, and at the same time, they improve security measures.

**2. Easy Integration:** As a matter of fact, it is easy to implement the use of your biometric facial systems in your company. They generally work well with the current software that you are already having.

**3. Automatic System:** Many businesses have shown that they value the automation of the biometric face-recognition system because they prefer the efficiency and the low labor requirements that these technologies come with. They will therefore not have to worry about physical presence, to be monitored, because the system is inherently an automation and as such, does not require it.

**4. Better Security**
Securements with a facial biometrics system will become significantly better. It is not only that you can keep a track of the employees but you can also add any visitor to the system and keep a track of them all over the premises. Any other person who is not registered in the system will not be allowed.

**5. Time Fraud not Exist**
Among the strongest benefits of using facial biometric systems in an association, it must be said that such a method helps to eliminate the time fraud issues. As a result, it is impossible to proxy associates since every member is required to present a facial scan in order to check in [50].

**8. Result**
Here, we provide an in-depth comparison of the results of the experiment presented in Table 4 that clarifies differences between databases, approaches, year of publishing, authors, and the findings.

**Table 4:** The Experimental Results

| No. | Author/s | Year | Technique | Database | Result |
| --- | --- | --- | --- | --- | --- |
| 1 | Vankayalapati et al. [51] | 2009 | CNN | ORL | 95% |
| 2 | Kong, Rui et al. [52] | 2011 | ICA, SVM | ORL | 96% |
| 3 | Bellakhdhar et al. [53] | 2013 | Magnitude Phase of Gabor, PCA, SVM | ORL, FRGCv2 | 99.90% |
| 4 | Jameel, S. [54] | 2015 | PCA + DCT | ORL | 95.122% |
| 5 | Fathima et al. [55] | 2015 | Gabor wavelet and linear discriminant analysis (LDA) | AT&T, MIT-India and Faces94 da- tasets | 88-94.02% |
| 6 | Ghorbel et al. [56] | 2016 | Eigenfaces and DoG filter | FERET | 84.26% |
| 7 | Bhaskar, A. et al. [57] | 2016 | SVM | Yale Faces | 97.78% |

| 8 | Fu *et al*. [58] | 2017 | Guided convolutional neural network, the loss function | CASIAWebface, LFW | 91.9-97.1% |
| 9 | Khan *et al*. [59] | 2018 | PCA | NCR-IIT facial database and real-time video stream | 69-86% |
| 10 | Banerjee *et al*. [60] | 2018 | Supervised learning, Viola-Jones, generic 3D model | PaSC videos and CW images, CMU multi-PIE dataset | 88.45-97.28% |

## 10. Applications of Face Recognition
The constant endeavor to maximize the accuracy of facial recognition systems is a result of the intensive applicability of the system in many fields. It has use in a variety of fields, including:

### 10.1. Law Enforce
The face recognition systems have been found to be very effective in the search of missing individuals and suspect identification by law-enforcement agencies. Hours of video-recordings are lengthy to manually search to recognize persons; it is a tedious process to law-enforcement administrators. Indicatively, the officials of the Chinese law-enforcement agencies have recently been reported to have identified BBC newsman John Sudworth in less than seven minutes using a powerful CCTV network that has more than 170 million cameras in addition to face-recognition capabilities. The recent innovations in the area of facial-recognition provided researchers with a new generation of smart and effective investigative tools of law-enforcement agencies. Another major challenge that arises before densely populated societies throughout the world is the issue of overstayed visas and unlawful residency.

Facial-recognition technologies have allowed the high-accuracy identification of individuals who have exceeded or otherwise breached immigration laws, thus allowing them to be arrested. Furthermore, the technology is being rolled out in different fields, such as immigration control, counter-terrorist actions, criminal investigations, voter registration, and financial services [61].

### 10.2. Entertainment
The spread of the facial-recognition technology is becoming more and more popular in the entertainment industry. Potential areas of development are theme-park gaming areas, training simulators, human-computer interaction, human-robot interaction, mobile gaming, and virtual-reality rooms, among others.

Furthermore, a new game has recently been presented by T. Feltwell *et al*. and it consists of the challenge that participants are presented with to differentiate between similarities in faces within an audience. Based on the example of freemium games like the iconic Pokémon game, the authors offer a variation experience where users actively hunt and catch real people [62].

### 10.3. Surveillance:
The surveillance is one of the most major and problematic projects in full automated and intelligent monitoring systems. In its traditional meaning, the term surveillance refers to the act of overseeing or keeping a close watch especially when criminal espionage is involved or when suspects are being pursued. It is one of the most popular and used applications in this field. The said systems are designed to meet the security goals related to both outdoor and indoor settings, which include geographical boundaries, bank halls, airports arrival bays, and the streets among others. As an overflow of data that camera networks can produce is enormous, the traditional brute force based detection algorithms cannot be effectively used to fulfill the advanced recognition of terrorists and suspects in sensitive areas. Recent developments in facialrecognition research provide a medium of intelligent video-monitoring systems, incorporating both software and hardware systems, including machine-learning systems, signal-processing systems, patternrecognition systems, and autonomous interfaces to produce highly optimistic outcomes (see reference) [63]. Secondly, machine-learning based methods of recognition have been proven to outperform human recognitions in a variety of real world scenarios, providing significant efficiency improvements. The face-recognition modules that are embedded in monitoring systems can become potent tools to help the personnel perform the recognition tasks and organize the process of complicated surveillance by leveraging the strength of both machines and people. However, with recent developments, although promising effects have been achieved, there are still significant challenges in efficient monitoring, including insufficient training cases, and subject images are also obstructed and blurred (see citation) [64].

### 10.4. Access Controlling
As the use of facial recognition has become more and more popular, such systems have been heavily integrated into automated access-control systems that control human-machinery interaction. As a result, other forms of authentication e.g. iris verification, biometrical fingerprints and cryptography passwords have been developed to complement or substitute these systems. The wide use of CCTV devices and cameras that are built-in smartphones has made facial recognition more achievable in a variety of viable conditions. Hardware validation Hardware based validation is also gaining fast, which allows a facial based authorization to single network log-in to multifaceted services. Consequently, facial biometric authentication on automated teller machines, ciphertext processing, and electronic transfer of money are becoming popular in various socio-technical environments [65].

### 10.5. Face Authentication in Mobile
Face authentication has also become a center of attention of mobile technology where an application can authenticate the identity of the user before accessing sensitive services like e-banking. We must urgently look into various elements of the face-recognition systems like its limits of deployment and susceptibility to presentation attacks, particularly within the mobile setting [41].

### 10.6. Another Common Applications
Recently Kwon and Lee [66] have described a wide range of face-recognition algorithms implanted in programs. Salici and Ciampini study the application of the facial-recognition systems to the forensic investigation in a complementary study and emphasize its practical applicability and methodological issues.

These techniques of identification are known to be successful in empirical assessments of 130 real world situations. The findings were supported by the forensic experts and thus, justified the accuracy and reliability of the methods that were used in the operations.

Besides that, Calo *et al*. introduce another modern implementation that aims at handling individual privacy over the lifecycle of electronic data. Their model deals with the issues of perception, updating, and safe deletion of information, thus presenting the wider discussion on the topic of privacy protection in the digital era.

**10.7 Face Recognition in Smart Cities**
The 2018 Market Research Report, states that the biometrics market is expected to grow by 13.89billion to 41.80billion between 2018 and 2023 and that the face-recognition technology is set to grow at a significantly higher rate during this time. In this regard, the facial-recognition market will be able to provide a high value in the new fields of application like identity verification in smart cities, smart homes, smart education, and, most importantly, e-government services, among others. In addition to technical issues, the usage of openly accessible facial-recognition systems can create ethical concerns and lead to the existing of the questions regarding the possibility of the usage and manipulation of the data which is one of the most sensitive kinds of privacy data. Such considerations are becoming recognized in the form of data-protection laws, most prominently, the General Data Protection Regulation (GDPR) of the EU, which aims at protecting people against unfair processing of their personal data by any third party. As a result, the necessity to protect the biometric data with the help of introducing new legal tools to regulate the use of the facial-recognition technologies is urgent.

**11. Conclusion**
The study of face recognition has been one of the difficult fields of scholars over the years. It has achieved a fair amount of publicity due to its numerous uses in practical fields namely computer graphics, pattern recognition, security and computer vision. Face recognition is a biometric technology that we present first in this survey. At this point we classify the face recognition models into three different groups, present the databases of faces used by researchers in this field to test their recognition methods and lastly give a summary of the experimental findings in Table 4. This paper will offer a broad overview of the major studies, which have been conducted on face recognition in different contexts. This literature review meant to help upcoming scholars in this field to come up with valuable methods, techniques, and models to enhance further research.

International Journal of Communication and Information Technology    http://www.computersciencejournals.com/ijcit